# African Trypanosomiasis Detection using Dempster-Shafer Theory

## Andino Maseleno, Md. Mahmud Hasan

Department of Computer Science, Faculty of Science, Universiti Brunei Darussalam
Jalan Tungku Link, Gadong BE 1410, Negara Brunei Darussalam
E-mail: andinomaseleno@yahoo.com, mahmud.hasan@ubd.edu.bn

## ABSTRACT

World Health Organization reports that African Trypanosomiasis affects mostly poor populations living in remote rural areas of Africa that can be fatal if properly not treated. This paper presents Dempster-Shafer Theory for the detection of African trypanosomiasis. Sustainable elimination of African trypanosomiasis as a public-health problem is feasible and requires continuous efforts and innovative approaches. In this research, we implement Dempster-Shafer theory for detecting African trypanosomiasis and displaying the result of detection process. We describe eleven symptoms as major symptoms which include fever, red urine, skin rash, paralysis, headache, bleeding around the bite, joint the paint, swollen lymph nodes, sleep disturbances, meningitis and arthritis. Dempster-Shafer theory to quantify the degree of belief, our approach uses Dempster-Shafer theory to combine beliefs under conditions of uncertainty and ignorance, and allows quantitative measurement of the belief and plausibility in our identification result.

**Keywords:** *African trypanosomiasis, detection, Dempster-Shafer theory*

## 1. INTRODUCTION

Human African Trypaniosomiasis or sleeping sickness is an old disease to be now considered as reemergent. Human African Trypaniosomiasis is endemic in 36 sub-Saharan African countries, in areas where tsetse flies are found [1]. The parasites concerned are protozoa belonging to the Trypanosoma genus. They are transmitted to humans by tsetse fly (Glossina genus) bites which have acquired their infection from human beings or from animals harbouring the human pathogenic parasites. Tsetse flies are found just in sub-Saharan Africa though only certain species transmit the disease. For reasons that are so far unexplained, there are many regions where tsetse flies are found, but sleeping sickness is not. Rural populations living in regions where transmission occurs and which depend on agriculture, fishing, animal husbandry or hunting are the most exposed to the tsetse fly and therefore to the disease. The disease develops in areas ranging from a single village to an entire region. Within an infected area, the intensity of the disease can vary from one village to the next. Between 2000 and 2009, out of 36 countries listed as endemic, 24 received the exclusive support of WHO (World Health Organization) either to assess the epidemiological status of HAT (human African trypanosomiasis) or to establish control and surveillance activities [2]. Another large group of vectors are flies. Sandfly species transmit the disease leishmaniasis, by acting as vectors for protozoan Leishmania species, and tsetse flies transmit protozoan trypansomes (Trypanosoma brucei gambiense and Trypansoma brucei rhodesiense) which cause African Trypanosomiasis (sleeping sickness). Ticks and lice form another large group of invertebrate vectors. The bacterium Borrelia burgdorferi, which causes Lyme Disease, is transmitted by ticks and members of the bacterial genus Rickettsia are transmitted by lice. For example, the human body louse transmits the bacterium Rickettsia prowazekii which causes epidemic typhus. Some systems for diagnosis in insect diseases have been developed which were expert system for identifies forest insects and proposes relevant treatment [3], and expert system of diseases and insects of jujube based on neural networks [4]. Actually, according to researchers knowledge, Dempster-Shafer theory of evidence has never been used for built an system for detecting African trypanosomiasis disease.

## 2. DEMPSTER-SHAFER THEORY

The Dempster-Shafer theory was first introduced by Dempster [5] and then extended by Shafer [6], but the kind of reasoning the theory uses can be found as far back as the seventeenth century. This theory is actually an extension to classic probabilistic uncertainty modeling. Whereas the Bayesian theory requires probabilities for each question of interest, belief functions allow us to base degrees of belief for on question on probabilities for a related question. Even though DST was not created specially in relation to artificial intelligence, the name Dempster-Shafer theory was coined by J. A. Barnett [7] in an article which marked the entry of the belief functions into the artificial intelligence literature. The Dempster-Shafer (D-S) theory or the theory of belief functions is a mathematical theory of evidence which can be interpreted as a generalization of probability theory in which the elements of the sample space to which nonzero probability mass is attributed are not single points but sets. The sets that get nonzero mass are called focal elements. The sum of these probability masses is one, however, the basic difference between D-S theory and traditional probability theory is that the focal elements of a Dempster-Shafer structure may overlap one another. The D-S theory also provides methods to represent and combine weights of evidence. m: $2^\Theta \to [0,1]$ is called a basic probability assignment (bpa) over $\Theta$ if it satisfies

m ($\emptyset$) = 0 and

$$\sum_{S \subset \emptyset} m(S) = 1 \qquad (1)$$

From the basic probability assignment, the upper and lower bounds of an interval can be defined. This interval contains the precise probability of a set of interest and is bounded by two nonadditive continuous measures called Belief (Bel) and Plausibility (Pl). The lower bound for a set $A$, Bel($A$) is defined as the sum of all the basic probability assignments of the proper subsets ($B$) of the set of interest ($A$) (B $\subseteq$ A ). Formally, for all sets $A$ that are elements of the power set, A $\in 2^\Theta$

$$\sum_{S \subset \emptyset} m(S) = 1 \qquad (2)$$

A function PL: $2^\Theta \to [0,1]$ is called a plausibility function satisfying

$$\sum_{B \cap A \neq \emptyset} m(B) \qquad (3)$$

The plausibility represents the upper bound for a set $A$, and is the sum of all the basic probability assignments of the sets ($B$) that intersect the set of interest ($A$) ( $B \cap A \neq \varphi$ ). The precise probability $P(A)$ of an event (in the classical sense) lies within the lower and upper bounds of Belief and Plausibility, respectively:

$$\text{Bel}(A) \le P(A) \le PL(A) \qquad (4)$$

The advantages of the Dempster-Shafer theory as follows:

1. It has the ability to model information in a flexible way without requiring a probability to be assigned to each element in a set,
2. It provides a convenient and simple mechanism (Dempster's combination rule) for combining two

or more pieces of evidence under certain conditions.
3. It can model ignorance explicitly.
4. Rejection of the law of additivity for belief in disjoint propositions.

Flowchart of African trypanosomiasis detection shown in Figure 1.

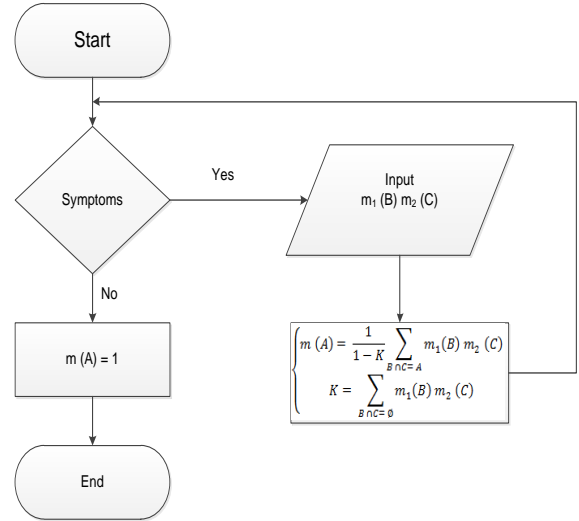

**Figure 1:** Flowchart of African Trypanosomiasis Detection

The consultation process begins with selecting the symptoms. If there are symptoms then will calculate, The Dempster-Shafer theory provides a rule to combine evidences from independent observers and into a single and more informative hint. Evidence theory is based on belief function and plausible reasoning. First of all, we must define a frame of discernment, indicated by the sign $\Theta$ . The sign $2^\Theta$ indicates the set composed of all the subset generated by the frame of discernment.

In this research, we used Dempster-Shafer theory for detecting African trypanosomiasis diseases. We describe eleven symptoms as major symptoms which include fever, red urine, skin rash, paralysis, headache, bleeding around the bite, joint the paint, swollen lymph nodes, sleep disturbances, meningitis and arthritis. Basic probability assignments for each symptom as shown in table 1. Suppose we have five different conditions

**Table 1:** Basic Probability Assignments for Each Symptom and Condition

| No. | Symptom | Diseases | Basic Probability Assignment | | | | |
|-----|---------|----------|-------------|-------------|-------------|-------------|-------------|
| | | | **Condition 1** | **Condition 2** | **Condition 3** | **Condition 4** | **Condition 5** |
| 1 | Fever | African trypanosomiasis | 0.65 | 0.65 | 0.65 | 0.65 | 0.45 |
| | | Babesiosis | | | | | |
| | | Dengue fever | | | | | |
| | | Malaria | | | | | |
| | | Rickettsial | | | | | |
| | | West nile | | | | | |
| 2 | Red urine | Babesiosis | 0.65 | 0.65 | 0.65 | 0.45 | 0.55 |
| 3 | Skin rash | Lyme | 0.65 | 0.65 | 0.45 | 0.55 | 0.45 |
| 4 | Paralysis | Lyme | 0.65 | 0.45 | 0.55 | 0.45 | 0.45 |
| 5 | Headache | Malaria | 0.45 | 0.55 | 0.45 | 0.45 | 0.55 |
| 6 | Bleeding around the bite | Rickettsial | 0.55 | 0.45 | 0.45 | 0.55 | 0.65 |
| 7 | Joint pain | African trypanosomiasis | 0.45 | 0.45 | 0.55 | 0.65 | 0.65 |
| 8 | Swollen lymph nodes | African trypanosomiasis | 0.45 | 0.55 | 0.65 | 0.65 | 0.65 |
| 9 | Sleep disturbances | African trypanosomiasis | 0.55 | 0.65 | 0.65 | 0.65 | 0.65 |
| 10 | Meningitis | West Nile | 0.65 | 0.65 | 0.65 | 0.65 | 0.65 |
| 11 | Arthritis | Dengue fever | 0.65 | 0.65 | 0.65 | 0.65 | 0.65 |

## 3. DETECTION OF AFRICAN TRYPANOSOMIASIS

The following will shown detection process of African trypanosomiasis {AT} using Dempster-Shafer Theory. Insect diseases which have similar symptoms with African trypanosomiasis include babesiosis {B}, dengue fever {DF}, malaria {M}, west nile {WN}, and lyme {L}.

1. Symptom 1: fever

Fever is a symptom of African trypanosomiasis, babesiosis, dengue fever, lyme, malaria, rickettsial, and west nile which has 0.65 for basic probability assignment.

$m_1$ {AT, B, DF, M, R, WN} = 0.65

$m_1$ { $\Theta$ } = 1 − 0.65 = 0.35     (5)

2. Symptoms 2: red urine

Red urine is a symptom of babesiosis which has 0.65 for basic probability assignment.

$m_2$ {B} = 0.65

$m_2$ { $\Theta$ } = 1 − 0.65 = 0.35     (6)

With red urine symptom then required to calculate the new bpa values for some combinations $(m_3)$. Combination rules for $m_3$ are shown in Table 2.

**Table 2:** The First Combination

| {B} | | **0.65** | $\Theta$ | **0.35** |
|-----|------|------|-----------------------|------|
| {AT, B, DF, M, R, WN} | 0.65 | {B} | 0.42 | {AT, B, DF, M, R, WN} | 0.23 |
| $\Theta$ | 0.35 | {B} | 0.23 | $\Theta$ | 0.12 |

$m_3$ {B} $= \dfrac{0.42 + 0.23}{1 - 0} = 0.65$

$m_3$ { AT, B, DF, M, R, WN }$= \dfrac{0.23}{1 - 0} = 0.23$

$m_3$ {$\Theta$} $= \dfrac{0.12}{1 - 0} = 0.12$     (7)

3. Symptom 3 : Skin rash

Skin rash is a symptom of lyme which has 0.65 for basic probability assignment.

$m_4$ {L} = 0.65

$m_4$ {$\Theta$} = 1 − 0.65 = 0.35     (8)

With skin rash symptom then required to calculate the new bpa values for some combinations $(m_5)$. Combination rules for $m_5$ are shown in Table 3.

**Table 3:** The Second Combination

| {L} | | **0.65** | $\Theta$ | **0.35** |
|-----|------|------|-----------------------|------|
| {B} | 0.65 | Ø | 0.42 | {B} | 0.23 |
| { AT, B, DF, M, R, WN } | 0.23 | Ø | 0.15 | { AT, B, DF, M, R, WN } | 0.08 |
| $\Theta$ | 0.12 | {L} | 0.08 | $\Theta$ | 0.04 |

$m_5$ {B} $= \dfrac{0.23}{1 - (0.42 + 0.15)} = 0.43$

$m_5$ {AT, B, DF, M, R, WN} $= \dfrac{0.08}{1 - (0.42 + 0.15)} = 0.19$

$$m_5 \{L\} = \frac{0.08}{1-(0.42+0.15)} = 0.19$$

$$m_5 \{\Theta\} = \frac{0.04}{1-(0.42+0.15)} = 0.09$$

(9)

### 4. Symptom 4: paralysis

Paralysis is a symptom of lyme which has 0.65 for basic probability assignment.

$m_6 \{L\} = 0.65$

$m_6 \{\Theta\} = 1 - 0.65 = 0.35$  (10)

With paralysis symptom then required to calculate the new bpa values for some combinations ($m_7$). Combination rules for $m_7$ are shown in Table 4.

**Table 4:** The Third Combination

| {L} | | **0.65** | | $\Theta$ | **0.35** |
|---|---|---|---|---|---|
| {B} | 0.43 | Ø | 0.28 | {B} | 0.15 |
| {AT, B, DF, M, R, WN} | 0.19 | Ø | 0.12 | {AT, B, DF, M, R, WN} | 0.07 |
| {L} | 0.19 | {L} | 0.12 | {L} | 0.07 |
| $\Theta$ | 0.09 | {L} | 0.06 | $\Theta$ | 0.03 |

$$m_7 \{B\} = \frac{0.15}{1-(0.28+0.12)} = 0.25$$

$$m_7 \{AT, B, DF, M, R, WN\} = \frac{0.07}{1-(0.28+0.12)} = 0.12$$

$$m_7 \{L\} = \frac{0.12+0.07+0.06}{1-(0.28+0.12)} = 0.42$$

$$m_7 \{\Theta\} = \frac{0.03}{1-(0.28+0.12)} = 0.05$$

(11)

### 5. Symptom 5: headache

Headache is a symptom of malaria which has 0.45 for basic probability assignment.

$m_8 \{M\} = 0.45$

$m_8 \{\Theta\} = 1 - 0.45 = 0.55$  (12)

With headache symptom then required to calculate the new bpa values for some combinations ($m_9$). Combination rules for $m_9$ are shown in Table 5.

**Table 5:** The Fourth Combination

| {M} | | **0.45** | | $\Theta$ | **0.55** |
|---|---|---|---|---|---|
| {B} | 0.25 | Ø | 0.11 | {B} | 0.14 |
| {AT, B, DF, M, R, WN} | 0.12 | {M} | 0.05 | {AT, B, DF, M, R, WN} | 0.07 |
| {L} | 0.42 | Ø | 0.19 | {L} | 0.23 |
| $\Theta$ | 0.05 | {M} | 0.02 | $\Theta$ | 0.03 |

$$m_9 \{B\} = \frac{0.14}{1-(0.11+0.19)} = 0.2$$

$$m_9 \{AT, B, DF, M, R, WN\} = \frac{0.07}{1-(0.11+0.19)} = 0.1$$

$$m_9 \{L\} = \frac{0.23}{1-(0.11+0.19)} = 0.33$$

$$m_9 \{M\} = \frac{0.05+0.02}{1-(0.11+0.19)} = 0.1$$

$$m_9 \{\Theta\} = \frac{0.03}{1-(0.11+0.19)} = 0.04$$

(13)

### 6. Symptom 6: bleeding around the bite

Bleeding around the bite is a symptom of rickettsial which has 0.55 for basic probability assignment.

$m_{10} \{R\} = 0.55$

$m_{10} \{\Theta\} = 1 - 0.55 = 0.45$  (14)

With bleeding around the bite symptom then required to calculate the new bpa values for some combinations ($m_{11}$). Combination rules for $m_{11}$ are shown in Table 6.

**Table 6:** The Fifth Combination

| {R} | | **0.55** | | $\Theta$ | **0.45** |
|---|---|---|---|---|---|
| {B} | 0.20 | Ø | 0.11 | {B} | 0.09 |
| {AT, B, DF, M, R, WN} | 0.10 | {R} | 0.05 | {AT, B, DF, M, R, WN} | 0.04 |
| {L} | 0.33 | Ø | 0.18 | {L} | 0.15 |
| {M} | 0.10 | Ø | 0.05 | {M} | 0.04 |
| $\Theta$ | 0.04 | {R} | 0.02 | $\Theta$ | 0.02 |

$$m_{11} \{B\} = \frac{0.09}{1-(0.11+0.18+0.05)} = 0.14$$

$$m_{11} \{AT, B, DF, M, R, WN\} = \frac{0.04}{1-(0.11+0.18+0.05)} = 0.06$$

$$m_{11} \{L\} = \frac{0.15}{1-(0.11+0.18+0.05)} = 0.23$$

$$m_{11} \{M\} = \frac{0.04}{1-(0.11+0.18+0.05)} = 0.06$$

$$m_{11} \{R\} = \frac{0.05+0.02}{1-(0.11+0.18+0.05)} = 0.11$$

$$m_{11} \{\Theta\} = \frac{0.02}{1-(0.11+0.18+0.05)} = 0.03$$

(15)

### 7. Symptom 7: joint the paint

Joint the paint is a symptom of African trypanosomiasis which has 0.45 for basic probability assignment.

$m_{12} \{AT\} = 0.45$

$m_{12} \{\Theta\} = 1 - 0.45 = 0.55$  (16)

With joint the paint symptom then required to calculate the new bpa values for some combinations ($m_{13}$). Combination rules for $m_{13}$ are shown in Table 7.

**Table 7:** The Sixth Combination

| {AT} | | **0.45** | Θ | | **0.55** |
|---|---|---|---|---|---|
| {B} | 0.14 | Ø | 0.06 | {B} | 0.08 |
| {AT, B, DF, M, R, WN} | 0.06 | {AT} | 0.03 | {AT, B, DF, M, R, WN} | 0.03 |
| {L} | 0.23 | Ø | 0.10 | {L} | 0.13 |
| {M} | 0.06 | Ø | 0.03 | {M} | 0.03 |
| {R} | 0.11 | Ø | 0.05 | {R} | 0.06 |
| Θ | 0.03 | {AT} | 0.01 | Θ | 0.02 |

$$m_{13}\{B\} = \frac{0.08}{1-(0.06+0.10+0.03+0.05)} = 0.11$$

$$m_{13}\{AT, B, DF, M, R, WN\} =$$
$$\frac{0.03}{1-(0.06+0.10+0.03+0.05)} = 0.04$$

$$m_{13}\{L\} = \frac{0.13}{1-(0.06+0.10+0.03+0.05)} = 0.17$$

$$m_{13}\{M\} = \frac{0.03}{1-(0.06+0.10+0.03+0.05)} = 0.04$$

$$m_{13}\{R\} = \frac{0.06}{1-(0.06+0.10+0.03+0.05)} = 0.08$$

$$m_{13}\{AT\} = \frac{0.03+0.01}{1-(0.06+0.10+0.03+0.05)} = 0.05$$

$$m_{13}\{\Theta\} = \frac{0.02}{1-(0.06+0.10+0.03+0.05)} = 0.03 \quad (17)$$

8.  Symptom 8: swollen lymph nodes

Swollen lymph nodes the paint is a symptom of African trypanosomiasis which has 0.45 for basic probability assignment.

$m_{14}\{AT\} = 0.45$

$m_{14}\{\Theta\} = 1 - 0.45 = 0.55 \quad (18)$

With swollen lymph nodes symptom then required to calculate the new bpa values for some combinations ($m_{17}$). Combination rules for $m_{17}$ are shown in Table 8.

**Table 8:** The Seventh Combination

| {AT} | | **0.45** | Θ | | **0.55** |
|---|---|---|---|---|---|
| {B} | 0.11 | Ø | 0.05 | {B} | 0.06 |
| {AT, B, DF, M, R, WN} | 0.04 | {AT} | 0.02 | {AT, B, DF, M, R, WN} | 0.02 |
| {L} | 0.17 | Ø | 0.08 | {L} | 0.09 |
| {M} | 0.04 | Ø | 0.02 | {M} | 0.02 |
| {R} | 0.08 | Ø | 0.04 | {R} | 0.04 |
| {AT} | 0.05 | {AT} | 0.02 | {AT} | 0.03 |
| Θ | 0.03 | {AT} | 0.01 | Θ | 0.02 |

$$m_{15}\{B\} = \frac{0.06}{1-(0.05+0.08+0.02+0.04)} = 0.07$$

$$m_{15}\{AT, B, DF, M, R, WN\} =$$
$$\frac{0.02}{1-(0.05+0.08+0.02+0.04)} = 0.02$$

$$m_{15}\{L\} = \frac{0.09}{1-(0.05+0.08+0.02+0.04)} = 0.11$$

$$m_{15}\{M\} = \frac{0.02}{1-(0.05+0.08+0.02+0.04)} = 0.02$$

$$m_{15}\{R\} = \frac{0.04}{1-(0.05+0.08+0.02+0.04)} = 0.05$$

$$m_{15}\{AT\} = \frac{0.02+0.02+0.03+0.01}{1-(0.05+0.08+0.02+0.04)} = 0.09$$

$$m_{15}\{\Theta\} = \frac{0.02}{1-(0.05+0.08+0.02+0.04)} = 0.02 \quad (19)$$

9.  Symptom 9: sleep disturbances

Sleep disturbances is a symptom of African trypanosomiasis which has 0.55 for basic probability assignment.

$m_{16}\{AT\} = 0.55$

$m_{16}\{\Theta\} = 1 - 0.55 = 0.45 \quad (20)$

With sleep disturbances symptom then required to calculate the new bpa values for some combinations ($m_{17}$). Combination rules for $m_{17}$ are shown in Table 9.

**Table 9:** The Nineth Combination

| {AT} | | **0.55** | Θ | | **0.45** |
|---|---|---|---|---|---|
| {B} | 0.07 | Ø | 0.04 | {B} | 0.03 |
| {AT, B, DF, M, R, WN} | 0.02 | {AT} | 0.01 | {AT, B, DF, M, R, WN} | 0.01 |
| {L} | 0.11 | Ø | 0.06 | {L} | 0.05 |
| {M} | 0.02 | Ø | 0.01 | {M} | 0.01 |
| {R} | 0.05 | Ø | 0.03 | {R} | 0.02 |
| {AT} | 0.09 | {AT} | 0.05 | {AT} | 0.04 |
| Θ | 0.02 | {AT} | 0.01 | Θ | 0.01 |

$$m_{17}\{B\} = \frac{0.03}{1-(0.04+0.06+0.01+0.03)} = 0.03$$

$$m_{17}\{AT, B, DF, M, R, WN\} =$$
$$\frac{0.01}{1-(0.04+0.06+0.01+0.03)} = 0.01$$

$$m_{17}\{L\} = \frac{0.05}{1-(0.04+0.06+0.01+0.03)} = 0.06$$

$$m_{17}\{M\} = \frac{0.01}{1-(0.04+0.06+0.01+0.03)} = 0.01$$

$$m_{17}\{R\} = \frac{0.02}{1-(0.04+0.06+0.01+0.03)} = 0.02$$

$$m_{17}\{AT\} = \frac{0.01+0.05+0.04+0.01}{1-(0.04+0.06+0.01+0.03)} = 0.13$$

$$m_{17}\{\Theta\} = \frac{0.01}{1-(0.04+0.06+0.01+0.03)} = 0.01 \quad (21)$$

## 10. Symptom 10: meningitis

Meningitis is a symptom of west nile which has 0.65 for basic probability assignment.

$m_{18}$ {WN} = 0.65

$m_{18}$ {Θ} = $1 - 0.65 = 0.35$  (22)

With meningitis symptom then required to calculate the new bpa values for some combinations ($m_{19}$). Combination rules for $m_{19}$ are shown in Table 10.

**Table 10:** The Tenth Combination

| {WN} | | **0.65** | Θ | **0.35** |
|---|---|---|---|---|
| {B} | 0.03 | Ø | 0.02 | {B} | 0.01 |
| {AT, B, DF, M, R, WN} | 0.01 | {WN} | 0.01 | {AT, B, DF, M, R, WN} | 0.003 |
| {L} | 0.06 | Ø | 0.04 | {L} | 0.02 |
| {M} | 0.01 | Ø | 0.01 | {M} | 0.003 |
| {R} | 0.02 | Ø | 0.01 | {R} | 0.01 |
| {AT} | 0.13 | Ø | 0.08 | {AT} | 0.05 |
| Θ | 0.01 | {WN} | 0.01 | Θ | 0.003 |

$m_{19}$ {B} = $\dfrac{0.01}{1-(0.02+0.04+0.01+0.01+0.08)} = 0.01$

$m_{19}$ {AT, B, DF, M, R, WN} =
$\dfrac{0.003}{1-(0.02+0.04+0.01+0.01+0.08)} = 0.003$

$m_{19}$ {L} = $\dfrac{0.02}{1-(0.02+0.04+0.01+0.01+0.08)} = 0.02$

$m_{19}$ {M} =
$\dfrac{0.003}{1-(0.02+0.04+0.01+0.01+0.08)} = 0.003$

$m_{19}$ {R} = $\dfrac{0.01}{1-(0.02+0.04+0.01+0.01+0.08)} = 0.01$

$m_{19}$ {AT} =
$\dfrac{0.05}{1-(0.02+0.04+0.01+0.01+0.08)} = 0.06$

$m_{19}$ {WN} =
$\dfrac{0.01+0.01}{1-(0.02+0.04+0.01+0.01+0.08)} = 0.02$

$m_{19}$ {Θ} =
$\dfrac{0.003}{1-(0.02+0.04+0.01+0.01+0.08)} = 0.003$  (23)

## 11. Symptom 11: Arthritis

Arthritis is a symptom of dengue fever which has 0.65 for basic probability assignment.

$m_{20}$ {DF} = 0.65

$m_{20}$ {Θ} = $1 - 0.65 = 0.35$  (24)

With arthritis symptom then required to calculate the new bpa values for some combinations ($m_{21}$). Combination rules for $m_{21}$ are shown in Table 11.

**Table 11:** The Eleventh Combination

| {DF} | | **0.65** | Θ | **0.35** |
|---|---|---|---|---|
| {B} | 0.01 | Ø | 0.01 | {B} | 0.003 |
| {AT, B, DF, M, R, WN} | 0.003 | {DF} | 0.001 | {AT, B, DF, M, R, WN} | 0.001 |
| {L} | 0.02 | Ø | 0.13 | {L} | 0.01 |
| {M} | 0.003 | Ø | 0.001 | {M} | 0.001 |
| {R} | 0.01 | Ø | 0.01 | {R} | 0.003 |
| {AT} | 0.06 | Ø | 0.04 | {AT} | 0.02 |
| {WN} | 0.02 | Ø | 0.13 | {WN} | 0.01 |
| Θ | 0.003 | {DF} | 0.001 | Θ | 0.001 |

$m_{21}$ {B} =
$\dfrac{0.003}{1-(0.01+0.13+0.001+0.01+0.04+0.13)} = 0.004$

$m_{21}$ {AT, B, DF, M, R, WN} =
$\dfrac{0.001}{1-(0.01+0.13+0.001+0.01+0.04+0.13)} = 0.001$

$m_{21}$ {L} =
$\dfrac{0.01}{1-(0.01+0.13+0.001+0.01+0.04+0.13)} = 0.01$

$m_{21}$ {M} =
$\dfrac{0.001}{1-(0.01+0.13+0.001+0.01+0.04+0.13)} = 0.001$

$m_{21}$ {R} =
$\dfrac{0.003}{1-(0.01+0.13+0.001+0.01+0.04+0.13)} = 0.004$

$m_{21}$ {AT} =
$\dfrac{0.02}{1-(0.01+0.13+0.001+0.01+0.04+0.13)} = 0.03$

$m_{21}$ {WN} =
$\dfrac{0.01}{1-(0.01+0.13+0.001+0.01+0.04+0.13)} = 0.01$

$m_{21}$ {DF} =
$\dfrac{0.001+0.001}{1-(0.01+0.13+0.001+0.01+0.04+0.13)} = 0.003$

$m_{21}$ {Θ} =
$\dfrac{0.001}{1-(0.01+0.13+0.001+0.01+0.04+0.13)} = 0.001$  (25)

We calculate other basic probability assignments of each symptom and condition, as is shown in equation (5) through (25).

## 4. RESULT

Figure 2 shows the highest basic probability assignment and the possibility of a temporary disease for each condition.

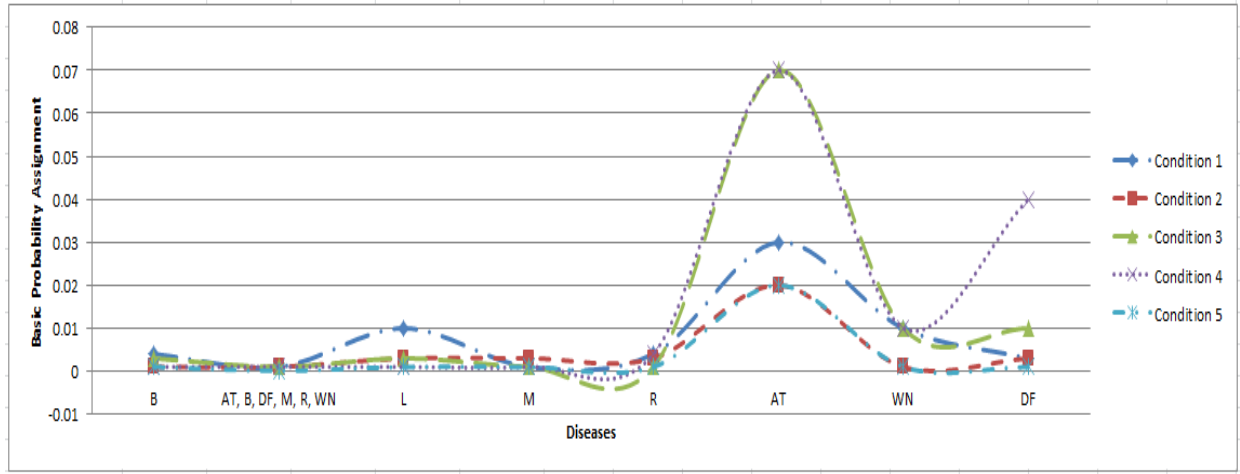

**Figure 2:** Final Result

The highest basic probability assignment and the possibility of a temporary disease for each condition is an African trypanosomiasis. Basis probability assignment for Condition 1 is 0.03, condition 2 is 0.02, condition 3 is 0.07, condition 4 is 0.07 and condition 5 is 0.02.

## 5.    CONCLUSION

In this paper, we described how the Dempster-Shafer theory can be used for dealing with African Trypanosomiasis detection. We described eleven symptoms as major symptoms which include fever, red urine, skin rash, paralysis, headache, bleeding around the bite, joint pain, swollen lymph nodes, sleep disturbances, meningitis, and arthritis. The simplest possible method for using probabilities to quantify the uncertainty in a database is that of attaching a probability to every member of a relation, and to use these values to provide the probability that a particular value is the correct answer to a particular query. The knowledge is uncertain in the collection of basic events can be directly used to draw conclusions in simple cases, however, in many cases the various events associated with each other. Knowledge base  is to draw conclusions, it is derived from uncertain knowledge. Reasoning under uncertainty that used some of mathematical expressions, gave them a different interpretation: each piece of evidence may support a subset containing several hypotheses. This is a generalization of the pure probabilistic framework in which every finding corresponds to a value of a variable.